%% file: COD.tex
\documentclass[lettersize,journal]{IEEEtran}
\usepackage{amsmath,amsfonts}
\usepackage{algorithmic}
\usepackage{algorithm}
\usepackage{array}
\usepackage{arydshln}
\usepackage{booktabs}
\usepackage[caption=false,font=normalsize,labelfont=sf,textfont=sf]{subfig}
\usepackage{cases}
\usepackage{color}
\usepackage{float}
\usepackage{makecell}
\usepackage{nicematrix}
\usepackage{multirow}
\usepackage{ragged2e}
\usepackage{subfloat}
\usepackage{stfloats}
\usepackage{textcomp}
\usepackage{url}
\usepackage{verbatim}
\usepackage{graphicx}
\usepackage{cite}

\usepackage[colorlinks,linkcolor=blue]{hyperref}

\input addon.tex

\usepackage{balance}

\begin{document}

\title{Frequency-Guided Spatial Adaptation for Camouflaged Object Detection}

\author{Shizhou Zhang*,
        Dexuan Kong*,
        Yinghui Xing$^{\dag}$,
        Yue Lu,
        Lingyan Ran,
        Guoqiang Liang,
        Hexu Wang,
        Yanning Zhang.
\thanks{This work was supported in part by the National Natural Science Foundation of China (NSFC) under Grant 62101453, 62201467; in part by the Young Talent Fund of Xi'an Association for Science and Technology under Grant 959202313088, in part by Innovation Capability Support Program of Shaanxi (Program No. 2024ZC-KJXX-043); in part by the Project funded by China Postdoctoral Science Foundation under Grant 2022TQ0260 and Grant 2023M742842; in part by the Fundamental Research Funds for the Central Universities No. HYGJZN202331 and in part by the Natural Science Basic Research Program of Shaanxi Province (No. 2022JC-DW-08).}
\thanks{Shizhou Zhang, Dexuan Kong, Yinghui Xing, Yue Lu, Lingyan Ran, Guoqiang Liang, and Yanning Zhang are with National Engineering Laboratory for Integrated Aero-Space-Ground-Ocean Big Data Application Technology, School of Computer Science, Northwestern Polytechnical University, Xi’an, China. Yinghui Xing is also with the Research \& Development Institute of Northwestern Polytechnical University in Shenzhen. Hexu Wang is with the School of Information and Technology, Northwest University, Xi'an 710127, P.R.China and Xi'an Key Laboratory of Human-Machine Integration and Control Technology for Intelligent Rehabilitation, Xijing University, Xi'an, 710123, Shaanxi, China.  
}
\thanks{*The first two authors equally contributed to this work. {\dag} correspondence author.} 
}

\maketitle

\begin{abstract}
Camouflaged object detection (COD) aims to segment camouflaged objects which exhibit very similar patterns with the surrounding environment. 
Recent research works have shown that enhancing the feature representation via the frequency information can greatly alleviate the ambiguity problem between the foreground objects and the background.
With the emergence of vision foundation models, like InternImage, Segment Anything Model etc, adapting the pretrained model on COD tasks with a lightweight adapter module shows a novel and promising research direction. 
Existing adapter modules mainly care about the feature adaptation in the spatial domain. 
In this paper, we propose a novel frequency-guided spatial adaptation method for COD task. 
Specifically, we transform the input features of the adapter into frequency domain. 
By grouping and interacting with frequency components located within non overlapping circles in the spectrogram, different frequency components are dynamically enhanced or weakened, making the intensity of image details and contour features adaptively adjusted. 
At the same time, the features that are conducive to distinguishing object and background are highlighted, indirectly implying the position and shape of camouflaged object. 
We conduct extensive experiments on four widely adopted benchmark datasets and the proposed method outperforms 26 state-of-the-art methods with large margins. Code will be released.
\end{abstract}

\begin{IEEEkeywords}
Camouflaged object detection, frequency-guided, pretrained foundation model, spatial adaptation.
\end{IEEEkeywords}

\section{Introduction}
 
\begin{figure}[ht!]
    \centering
    \begin{center}
       \includegraphics[width=1\linewidth, height=0.70\linewidth]{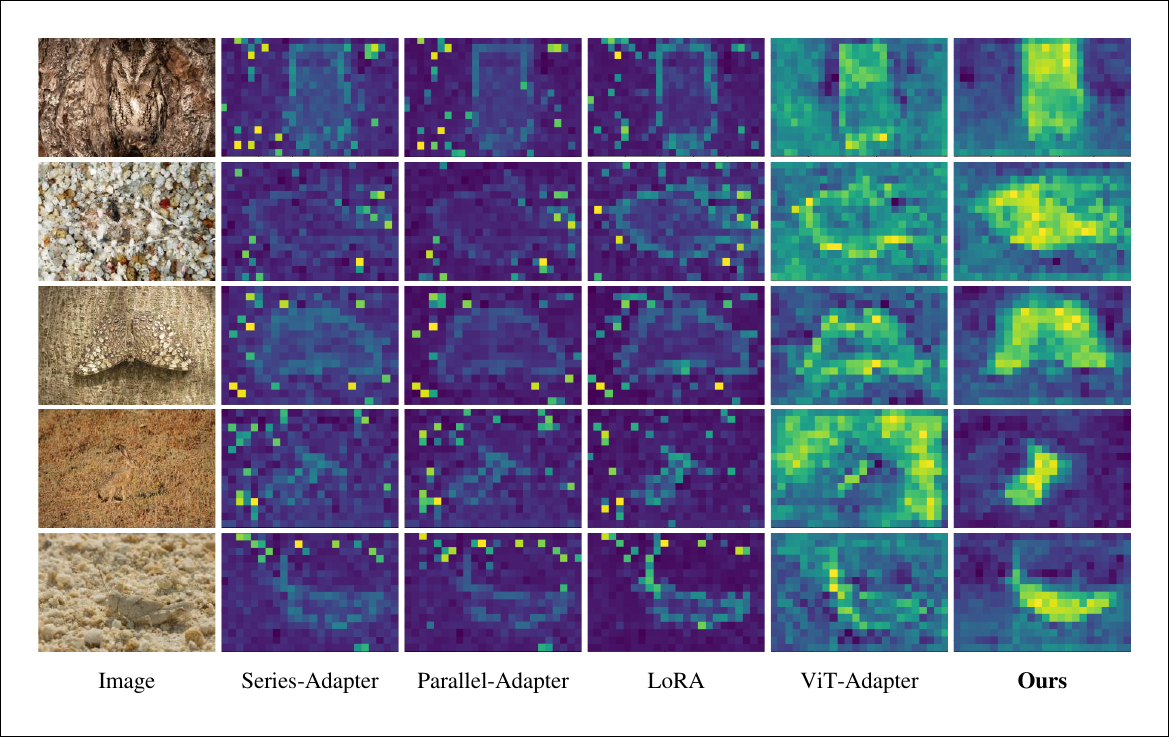}
    \end{center}
    \caption{Visualization and comparison of feature maps obtained after adapter tuning on COD task.}
    \label{fig:fig1}
\end{figure}
\IEEEPARstart{C}{amouflaged} object detection (COD), which has wide downstream applications such as medical segmentation~\cite{fan2020pranet,huang2020mcmt} and recreational art~\cite{dean2017art}, aims to segment the objects which are perfectly embedded in their surrounding environment.
Recent years have witnessed the great progress of COD, it remains a challenging task due to the low contrast appearances between the concealed objects and the background. 
In addition, the semantic categories of the objects lie between a wide range from naturally camouflaged objects such as mammals or insects hiding themselves from their predators, to artificially camouflaged objects such as soldiers on the battlefields or human body painting arts.
The diverse types of objects with various shapes, sizes and textures further increases the difficulties of the COD task.

From one hand, some recent methods try to design progressively coarse to fine feature enhancement process~\cite{ConcealedObject2022,HighResolutionIterative2023} or to utilize extra edge information~\cite{MutualGraph2021} to locate accurate boundaries from the spatial/RGB domain information of an image. 
While other works propose to introduce clues in frequency domain~\cite{DetectingCamouflaged2022,FrequencyPerception2023,FrequencyawareCamouflaged2023},  as the frequency enhanced features are more discriminative between the concealed objects and background.

From the other hand, with the emergence of large scale pretrained vision foundation models, such as InternImage~\cite{wang2023internimage} and Segment Anything Model (SAM)~\cite{kirillov2023segany}, a promising research paradigm which is prevalent on almost all vision tasks is that adapting the foundation model on the downstream tasks with a small portion of extra trainable parameters or architectures, e.g. prompts and adapters, while the parameters of the pretrained model kept frozen. 
Existing task-specific adapters broadly fall into three categories: series adapter~\cite{hu2023llm}, parallel adapter~\cite{hu2023llm} and LoRA~\cite{LoRALowRank2021}. 
To introduce the image-related inductive biases into the pretrained ViT model for pixel-wise dense prediction tasks, ~\cite{chen2022vision} proposed a specific parallel ViT-Adapter to further aggregate multi-scale context.
Current adapters are devised to compensate the features or weights all from the spatial domain. 
However, crucial clues for the downstream COD task, such as subtle variations in textures and patterns, may not be easily observed in the spatial domain but can be revealed by the unique spectral characteristics in the frequency domain. Therefore, 
adapting the pretrained foundation model from the spatial domain alone can not take the full advantage of the merits brought by the frequency domain information which is especially required for the COD task.

In this paper, we propose a novel adaptation model named as frequency-guided spatial adaptation network (FGSA-Net) for COD task.
Firstly, we devise a frequency-guided spatial attention (FGSAttn) module by transforming the input features of the adapter into frequency domain. 
Then by grouping and interacting with frequency components located within non overlapping circles in the spectrogram, different frequency components are enhanced or weakened, making the intensity of image details and contour features adaptively adjusted.

Based on the FGSAttn module and the multi-scale context aggregation as in ViT-Adapter, we further propose a Frequency-Based Nuances Mining (FBNM) module which aims to mining subtle differences between foreground and background, and a Frequency-Based Feature Enhancement (FBFE) module which extracts and fuses multi-scale features containing general knowledge of the pretrained model and adaptation components learned from the new data of downstream COD task. 
As can be seen from Fig.~\ref{fig:Framework}, the FBNM module is inserted after the patch embedding layer and the FBFE module is inserted into the pretrained ViT backbone model after each $K$ layers.
During training, only the parameters of FBNM and FBFE modules are optimized while the parameters of pretrained ViT model are kept frozen. 
{With only about 7\% tunable parameters (over the total parameters of the pretrained model)}, our proposed FGSA-Net achieves state-of-the-art performances on four widely adopted benchmark datasets of COD and outperforms the spatial adaptation counterparts with a large margin. 
Fig.~\ref{fig:fig1} illustrates the obtained feature maps after adaptation, it can be seen that our proposed novel adaptation mechanism clearly concentrate more on the concealed objects compared with other spatial adaptation methods.


To summarize, the contributions of this paper are threefolds,
\begin{itemize}
    \item We propose a novel frequency-guided spatial adaptation network, which combines the advantage of general knowledge of vision foundation model and task-specific features learned from the new data of downstream COD task.
    \item A frequency-guided spatial attention module is devised to adapt the pretrained foundation model from spatial domain while guided by the adaptively adjusted frequency components to focus more on the camouflaged regions.
    \item The proposed method greatly outperforms the baseline methods and achieves state-of-the-art performances on four widely adopted COD benchmark datasets.
\end{itemize}

The rest of the paper is organized as follows, Section~\ref{sec_RW} reviews the relevant works of our paper. Section~\ref{sec_method} elaborates each component of the proposed method. In Section~\ref{sec_expr}, we conduct thorough experiments on four widely-used benchmark datasets to verify the superiority of our method and further analyze the effectiveness of each component. Finally, we draw the conclusion of the paper in Section~\ref{sec_conclusion}.

\section{Related Work}\label{sec_RW}
\subsection{Camouflaged Object Detection}
Numerous efforts have been undertaken in the field of camouflaged object detection~\cite{HighResolutionIterative2023,BoundaryGuidedCamouflaged2022,SegmentMagnify2022,ConcealedObject2022,InferringCamouflaged2021,ContextawareCrosslevel2021,MutualGraph2021,UncertaintyGuidedTransformer2021,UncertaintyAwareJoint2021,lyu2023uedg,zhai2022deep,xing4790957multi}. 
In order to obtain accurate boundary,~\cite{BoundaryGuidedCamouflaged2022} promote the model to generate features that highlight object structure for accurate boundary localization of camouflaged objects.~\cite{MutualGraph2021} decouple an image into two feature maps and recurrently reason their high-order relations through graphs for roughly locating the target and accurately capturing its boundary details.~\cite{lyu2023uedg} combine probabilistic-derived uncertainty
and deterministic-derived edge information to accurately detect concealed objects.
To capture rich features of camouflaged objects,~\cite{ContextawareCrosslevel2021} integrate and fuse multi-level image features to yield multi-scale representations for exploiting rich global context information.~\cite{HighResolutionIterative2023} iteratively refine low-resolution representations by high-resolution features to extract high-resolution texture details and avoid the detail degradation.~\cite{zhai2022deep} leverage the spatial organization of textons in the foreground and background regions as discriminative cues for camouflaged object detection.
Additionally, some recent works~\cite{DetectingCamouflaged2022,FrequencyPerception2023} investigate that clues in frequency domain can help the feature enhancement of concealed objects.

\subsection{Parameter-Efficient Fine-Tuning}
Parameter-efficient fine-tuning aims to adapt pretrained models to downstream tasks by inserting a few learnable parameters~\cite{VisualPrompt2022,TipAdapterTrainingfree2021,LoRALowRank2021,UnifiedView2021,ParameterEfficientTransfer2020,ParameterEfficientTransfer2019,xing2023dual}.
~\cite{ParameterEfficientTransfer2019} propose lightweight adapters for Transformers \cite{AttentionAll2017a} in the filed of NLP.
~\cite{VisualPrompt2022} introduce learnable tokens (\textit{i.e.} prompts) into Vision Transformers~\cite{ImageWorth2020}.
~\cite{LoRALowRank2021} inject trainable rank decomposition matrices into each layer of the Transformer architecture.
All the methods only update the introduced small number of parameters (\textit{i.e.} adapters, prompts, \textit{etc.}) while keep the pretrained parameters fixed.
Consequently, the training process requires much less memory and computation costs than fine-tuning the whole model. 
However, existing adapters deal with the feature adaptation problem from the spatial domain alone.

\subsection{Frequency-Based Methods}
Since the features of camouflaged objects and the background are more discriminative in the frequency domain, a line of approaches~\cite{FourmerEfficient2023,FrequencyawareCamouflaged2023,EmbeddingFourier2023,FrequencyPerception2023,xing2022learning,DetectingCamouflaged2022,FcaNetFrequency2021,LearningFrequency2020,xing2023pansharpening} dig frequency clues for camouflaged object detection or other tasks to enhance the feature representation.
~\cite{DetectingCamouflaged2022} adopt the offline discrete cosine transform to extract frequency features, and then fuse the features from RGB domain and frequency domain.
~\cite{FrequencyawareCamouflaged2023} aggregate multi-scale features from a frequency perspective and enhance the features of the learned important frequency components.
~\cite{FrequencyPerception2023} utilize the octave convolution ~\cite{DropOctave2019} in the frequency perception module for coarse positioning, and combine high-level features with shallow features to achieve the detailed correction of the camouflaged objects.
Different from the attention modules only performed in RGB domain, we exploit the spatial adapter while guided with frequency domain information, which is more helpful to distinguish between camouflaged objects and the background.


\begin{figure*}[ht!]
    \centering
    \resizebox{\linewidth}{!}{
    \belowrulesep=0pt
    \aboverulesep=0pt
    \includegraphics{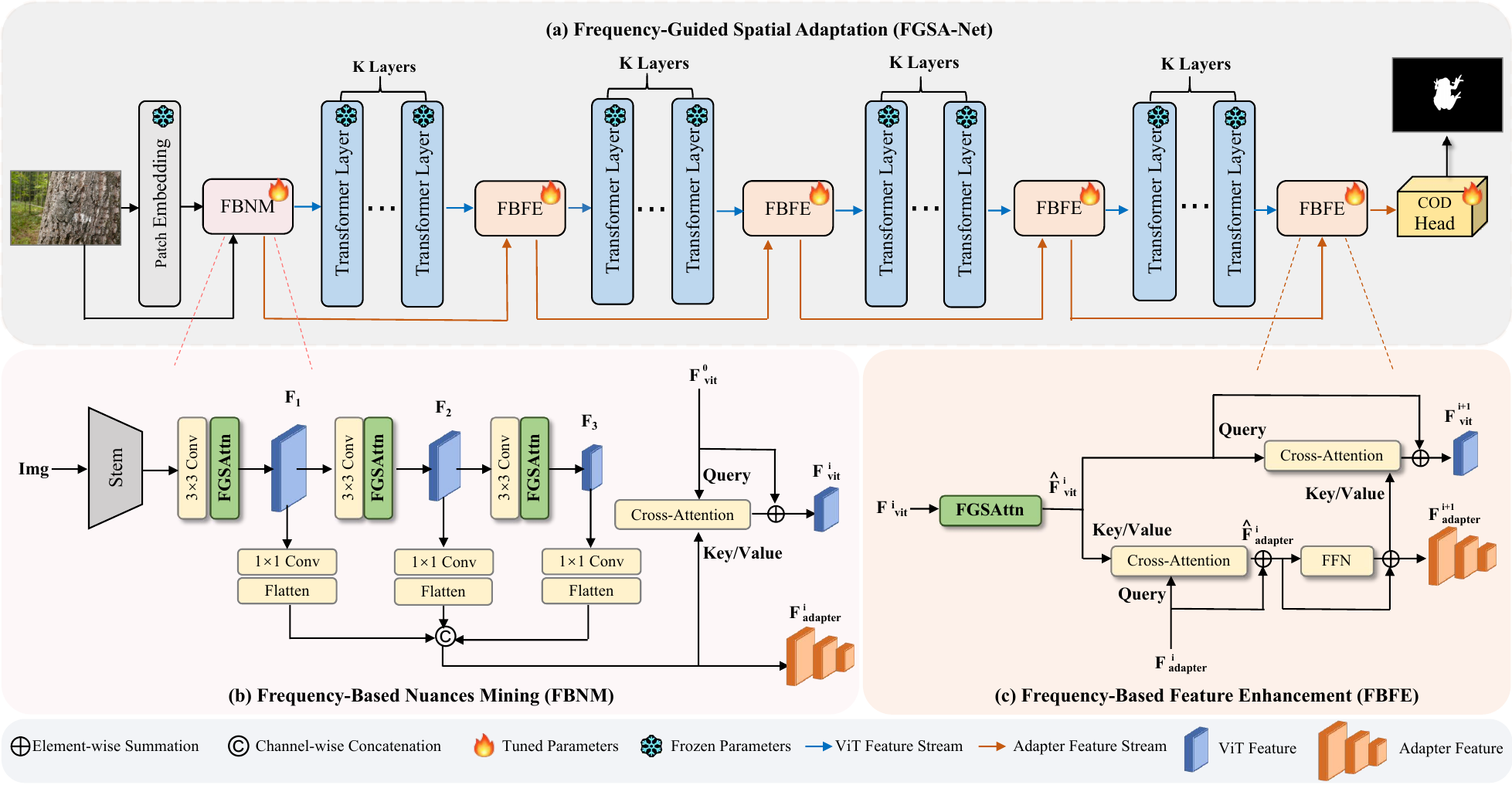}
    }
    \caption{Overall framework of our proposed FGSA-Net. (a) The main architecture. (b) Frequency-Based Nuances Mining module. (c) Frequency-Based Feature Enhancement module. }
    \label{fig:Framework}
\end{figure*}

\section{Methodology}\label{sec_method}
\subsection{Overview}


%
{As a typical low contrast structural segmentation task, COD methods require not only low-level structural details but also global context information. 
{However, available adaptation models like ViT-adapter~\cite{chen2022vision} and SAM-adapter~\cite{chen2023sam}, only consider global context information in spatial domain, limiting their ability to locate subtle differences between foreground and background.} 
To accurately represent refined structure of camouflaged objects, we resort to extract and enhance detail information from the frequency perspective to design a frequency-guided spatial adapter (FGSA-Net). 
{The overall architecture is shown in Fig.~\ref{fig:Framework} (a), 
including a large pretrained ViT model, a lightweight adapter module consist of frequency-based nuances mining (FBNM) and frequency-based feature enhancement (FBFE), as well as a detection head for COD.
Specifically, as shown in Fig.~\ref{fig:FGSAttn}, we devise the frequency-guided spatial attention (FGSAttn) module to concentrate more on the concealed objects by dynamically adjusting the frequency components.
Based on the FGSAttn, as detailed in Fig.~\ref{fig:Framework} (b) and Fig.~\ref{fig:Framework} (c), two elaborate modules, i.e., FBNM and FBFE, responsible for subtle feature extraction and enhancement, are proposed to serve as the adapter. 
Among them, the FBNM module aims to receive original input images and serialized tokens to capture prior knowledge of the subtle differences between the foreground and background.
Then, we evenly split the transformer layers of ViT model into M groups, each of which contains $K$ layers.
The FBFE module is inserted into the ViT model after each group and performs interaction operations on the general knowledge from the pretrained ViT branch and the task-specific knowledge from the adapter branch to recalibrate the feature distribution. }
Finally, the hierarchical features output by the last FBFE module are input into the detection head to generate more refined and accurate prediction map.

\subsection{Frequency-Guided Spatial Attention Module}

\begin{figure*}[ht!]
    \centering
    \resizebox{\linewidth}{!}{
    \belowrulesep=0pt
    \aboverulesep=0pt
    \includegraphics{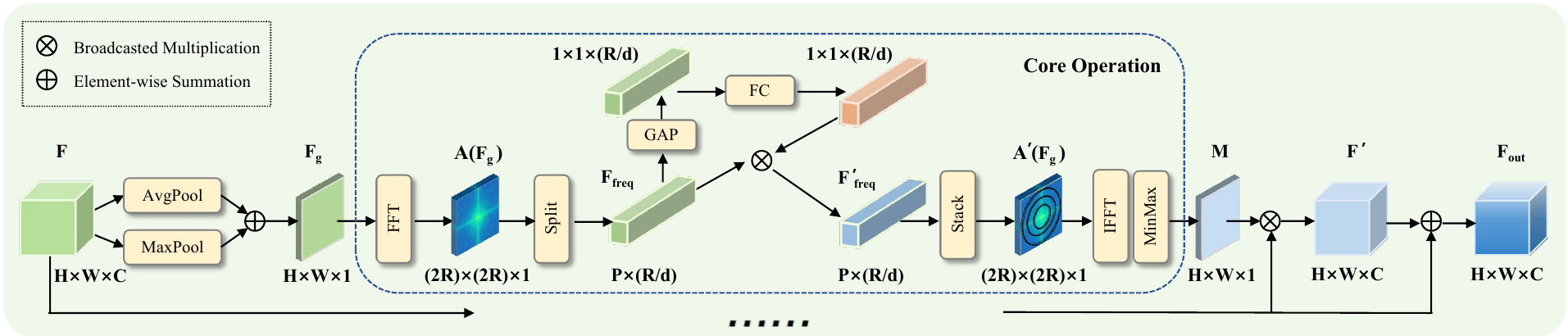}
    }
    \caption{The detailed architecture of our proposed Frequency-Guided Spatial Attention (FGSAttn) Module.}
    \label{fig:FGSAttn}
\end{figure*}
The detailed architecture of frequency-guided spatial attention (FGSAttn) module is 
illustrated in Fig.~\ref{fig:FGSAttn}. 
The input feature $\vF \in \mathbb{R}^{H \times W \times C}$ are processed by
average-pooling and max-pooling operations along the channel dimension, and the pooled features are then combined through an element-wise summation to 
obtain a single channel most distinctive features
$\vF_{g} \in \mathbb{R}^{H \times W \times 1}$. It can be formulated as:
\begin{equation}
\vF_g = AvgPool(\vF) + MaxPool(\vF).
\label{eq1}
\end{equation}

Then, $\vF_g$ is transformed into the frequency domain by Fourier transform 
to obtain its amplitude spectrum $\mathcal{A}(\vF_g) \in \mathbb{R}^{(2R) \times (2R) \times 1}$ and the phase spectrum $\mathcal{P}(\vF_g) \in \mathbb{R}^{(2R) \times (2R) \times 1}$, 
\begin{align}\label{eq2}
\mathcal{FT}(\vF_g)(u,v) &= \frac{1}{\sqrt{HW}}\sum_{h=0}^{H-1}\sum_{w=0}^{W-1}\vF_g(h,w)e^{-j2\pi(\frac{h}{H} u + \frac{w}{W} v)},\\ \nonumber
\mathcal{A}(\vF_g)(u, v) &= \sqrt{R^2(\vF_g)(u, v) + I^2(\vF_g)(u, v)},\\
\mathcal{P}(\vF_g)(u, v) &= \arctan{[\frac{I(\vF_g)(u, v)}{R(\vF_g)(u, v)}]}, \nonumber
\end{align}
where $\mathcal{FT}(.)$ represents the fast Fourier transform of the feature. $R(\vF_g)$ and $I(\vF_g)$ are the real and imaginary part of $\mathcal{FT}(\vF_g)$, respectively.

Previous studies~\cite{FourmerEfficient2023,zhang2024frequency} have proved that the amplitude component obtained by the Fourier transform contains more critical information for the object. 
Hence, in our work, we mainly explore the influence of different frequency components in the amplitude spectrum while keeping the phase spectrum unchanged.

After frequency centralization, the origin point in the amplitude spectrum represents center frequency. 
The distances between a certain point and the origin denotes its frequency component. Therefore, circles with different diameters in the amplitude spectrum correspond to different frequency components. 
When they are transformed back to the spatial domain, they represent different type of features, 
such as the approximate shape or edge details of objects. 
In our method, we decompose the amplitude spectrum into many non-overlapping circular rings with a width of $d$ along the radius dimension, 
{and the hyperparameter $d$ defines the range of frequency components.}

We group the features located in the same circular ring into one channel and obtain $\vF_{freq} \in \mathbb{R}^{P \times (R/d))}$. 
{$P$ denotes the number of frequency components on each channel.}
Then the $GAP(.)$ and $FC(.)$ are performed to generate weights for adaptively recalibrating the responses of
different frequency components.
Thus, mapping the adjusted frequency components back into the spatial domain 
will change the spatial feature values on the feature map, i.e. frequency-guided spatial adaptation, which can be expressed as:
\begin{equation}
\vF^{'}_{freq} = \vF_{freq} \otimes FC(GAP(\vF_{freq}))
\label{eq5}
\end{equation}
where ``$\otimes$" denotes element-wise broadcasted multiplication.
$GAP(.)$ and $FC(.)$ represent global average pooling and sequences of $1 \times 1$ convolutions followed by a LeakyReLU activation function. 

Finally, the processed features
$\vF^{'}_{freq}$ are re-arranged and stacked sequentially 
to get a new amplitude spectrum $\mathcal{A}^{'}(\vF_{g})$. 
Combined with original phase spectrum $\mathcal{P}(\vF_g)$, the spatial attention map $\vM \in \mathbb{R}^{H \times W \times 1}$ is then obtained by inverse Fourier transform,
\begin{equation}
\vM = MinMax(\mathcal{FT}^{-1}(\mathcal{A}^{'}(\vF_{g}), \mathcal{P}(\vF_g))).
\label{eq6}
\end{equation}
The final output $\vF_{out}$ is
\begin{equation}
\vF_{out} = \vF + \vM \otimes \vF,
\label{eq7}
\end{equation}
where $\mathcal{FT}^{-1}$ represents the inverse Fourier transform. 
``$\otimes$" denotes the element-wise broadcasted multiplication along the channel dimension.

\subsection{Frequency-Based Nuances Mining Module}
Since camouflaged objects always exhibit very similar appearance features with nearby noisy objects and background, the slight differences are difficult to be distinguished by the spatial domain features of the foundation model alone. 
We design a Frequency-Based Nuances Mining (FBNM) module aiming at mining nuances between foreground and background, and the detailed architecture is shown in  Fig.~\ref{fig:Framework}(b).

Specifically, a standard convolution stem borrowed from ResNet is employed to model the local spatial contexts of the input image, which consists of three convolutions and a max-pooling layer.
{After that, three consecutive sequences are applied to gradually aggregate multi-scale features with three resolutions of 1/8, 1/16, and 1/32, obtaining a feature pyramid of similar resolutions to FPN~\cite{lin2017feature}, which is widely used in dense prediction tasks.}
Each sequence contains a 3$\times$3 convolution kernel to reduce the scale of the feature map, followed by a FGSAttn module which leverages the frequency components to adjust feature layers representing different visual attributes from a global perspective. 
This can effectively highlight the nuance parts in texture-rich regions to distinguish the foreground and background.

Next, we project the feature maps to the same dimension $D$ using several 1x1 convolution layers. 
After a flatten layer and a concatenate layer, a feature pyramid $\vF^{i}_{adapter} \in \mathbb{R}^{(\frac{HW}{8^2}+\frac{HW}{16^2}+\frac{HW}{32^2}) \times D}$ can be then obtained. 
On one hand, it serves as the input for the next adapter module. 
On the other hand, it is injected into serialized tokens $F^0_{vit}$ via cross-attention mechanism to obtain $F^i_{vit}$ that absorb task related knowledge, which will be used as the input for the successive layers of the pretrained ViT backbone.
\begin{table*}[ht!]
    \caption{Quantitative comparison of our FGSA-Net and 26 SOTA methods for COD on four benchmark datasets. $\uparrow$ / $\downarrow$ indicates that larger/smaller is better. The best and second best are \textbf{bolded} and \underline{underlined} for highlighting, respectively. “—”: Not available.}
    \centering
    \renewcommand\arraystretch{1.3}
    \resizebox{\linewidth}{!}{
    \belowrulesep=3pt
    \aboverulesep=3pt
    \begin{tabular}{cc*{4}{c}*{4}{c}*{4}{c}*{4}{c}}
    \toprule
        \multirow{2}*{\textbf{Methods}} &\multirow{2}*{\textbf{Pub/Year}} & \multicolumn{4}{c}{\textbf{CHAMELEON (76)}} & \multicolumn{4}{c}{\textbf{CAMO-Test (250)}} & \multicolumn{4}{c}{\textbf{COD10K-Test (2,026)}} & \multicolumn{4}{c}{\textbf{NC4K-Test (4,121)}} \\ 
        \cmidrule(lr){3-6}\cmidrule(lr){7-10}\cmidrule(lr){11-14}\cmidrule(lr){15-18}
         ~ & ~ & 
         $S_{\alpha}\uparrow$ & $E_{\phi}\uparrow$ & $F^w_{\beta}\uparrow$ & 
         $~M~\downarrow$ & 
         $S_{\alpha}\uparrow$ & $E_{\phi}\uparrow$ & $F^w_{\beta}\uparrow$ & 
         $~M~\downarrow$ &
         $S_{\alpha}\uparrow$ & $E_{\phi}\uparrow$ & $F^w_{\beta}\uparrow$ & 
         $~M~\downarrow$ &
         $S_{\alpha}\uparrow$ & $E_{\phi}\uparrow$ & $F^w_{\beta}\uparrow$ & 
         $~M~\downarrow$\\
         
         \cmidrule(lr){1-18} 
        SINet & CVPR\_{20} & .869 & .891 & .740 & .044 & .751 & .771 & .606 & .100 & .771 & .806 & .551 & .051 & .808 & .871 & .723 & .058 \\ 
        PraNet & MICCAI\_{20} & .860 & .898 & .763 & .044 & .769 & .824 & .663 & .094 & .789 & .861 & .629 & .045 & .822 & .876 & .724 & .059 \\
        TINet &  AAAI\_{21} & .874 & .916 & .783 & .038 & .781 & .847 & .678 & .087 & .793 & .848 & .635 & .043 & .829 & .879 & .734 & .055 \\ 
        PFNet & CVPR\_{21} & .882 & .942 & .810 & .033 & .782 & .852 & .695 & .085 & .800 & .868 & .660 & .040 & .829 & .888 & .745 & .053 \\ 
        UGTR & ICCV\_{21} & .888 & .918 & .796 & .031 & .785 & .859 & .686 & .086 & .818 & .850 & .667 & .035 & .839 & .874 & .747 & .052 \\  
        C$^2$FNet & IJCAI\_{21} & .888 & .935 & .828 & .032 & .796 & .854 & .719 & .080 & .813 & .890 & .686 & .036 & .838 & .897 & .762 & .049 \\        
        S-MGL & CVPR\_{21} & .892 & .921 & .803 & .032 & .772 & .850 & .664 & .089 & .811 & .851 & .655 & .037 & .829 & .863 & .731 & .055 \\ 
        R-MGL & CVPR\_{21} & .893 & .923 & .813 & .030 & .775 & .847 & .673 & .088 & .814 & .865 & .666 & .035 & .833 & .867 & .740 & .052 \\ 
        LSR & CVPR\_{21} & .893 & .938 & .839 & .033 & .793 & .826 & .725 & .085 & .793 & .868 & .685 & .041 & .839 & .883 & .779 & .053 \\  
        JCSOD & CVPR\_{21} & .894 & .943 & .848 & .030 & .803 & .853 & .759 & .076 & .817 & .892 & .726 & .035 & .842 & .898 & .771 & .047 \\  
        ERRNet & PR\_{22} & .877 & .927 & .805 & .036 & .761 & .817 & .660 & .088 & .780 & .867 & .629 & .044 & .787 & .848 & .638 & .070 \\
        BASNet & AAAI\_{22} & .914 & .954 & .866 & .022 & .749 & .796 & .646 & .096 & .802 & .855 & .677 & .038 & .817 & .859 & .732 & .058 \\
        SINetV2 & TPAMI\_{22} & .888 & .942 & .816 & .030 & .820 & .882 & .743 & .070 & .815 & .887 & .680 & .037 & .847 & .903 & .770 & .048 \\          
        ZoomNet & CVPR\_{22} & .902 & .958 & .845 & .023 & .820 & .892 & .752 & .066 & .838 & .911 & .729 & .029 & .853 & .912 & .784 & .043 \\ 
        PENet & IJCAI\_{23} & .902 & .960 & .851 & .024 & .828 & .890 & .771 & .063 & .831 & .908 & .723 & .031 & .855 & .912 & .795 & .042 \\
        MFFN & WACV\_{23} & .905 & .963 & .852 & .021 & - & - & - & - & .846 & .917 & .745 & .028 & .856 & .915 & .791 & .042 \\
        FSPNet & CVPR\_{23} & - & - & - & - & .856 & .899 & .799 & .050 & .851 & .895 & .735 & .026 & .879 & .915 & .816 & \underline{.035} \\
        
        DINet & TMM\_{24} & - & - & - & - & .821 & .874 & .790 & .068 & .832 & .903 & .761 & .031 & .856 & .909 & .824 & .043 \\
        DTC-Net & TMM\_{22} & .876 & .897 & .773 & .039 & .778 & .804 & .667 & .084 & .790 & .821 & .616 & .041 & - & - & - & - \\
        HitNet & AAAI\_{23} & \textbf{.922} & \underline{.970} & \textbf{.903} & \underline{.018} & .844 & .902 & .801 & .057 & .868 & \underline{.932} & .798 & \underline{.024} & .870 & .921 & .825 & .039 \\
        
        UEDG & TMM\_{23} & .911 & .960 & .866 & .022 & \underline{.868} & \underline{.922} & \underline{.819} & \underline{.048} & .858 & .924 & .766 & .025 & \underline{.881} & \underline{.928} & \underline{.829} & \underline{.035} \\
        
        \cmidrule(lr){1-18}
        FDNet & CVPR\_{22} & .898 & .949 & .837 & .027 & .844 & .898 & .778 & .062 & .837 & .918 & .731 & .030 & - & - & - & - \\
        FBNet & MCCA\_{23} & .888 & .939 & .828 & .032 & .783 & .839 & .702 & .081 & .809 & .889 & .684 & .035 & - & - & - & - \\
        FPNet & MM\_{23} & .914 & .961 & .85 & .022 & .852 & .905 & .806 & .056 & .850 & .913 & .748 & .029 & - & - & - & - \\
        FEDER-MS-4 & CVPR\_{23} & .907 & .964 & .874 & .025 & .822 & .886 & .809 & .067 & .851 & .917 & .752 & .028 & .863 & .917 & .827 & .042 \\
        
        \cmidrule(lr){1-18}
        SAM-Adapter & ICCVW\_{23} & .896 & .919 & .824 & .033 & .847 & .873 & .765 & .070 & \underline{.883} & .918 & \underline{.801} & .025 & - & - & - & - \\
        
        \cmidrule(lr){1-18}
        \textbf{Ours(FGSA-Net)} & - & {\underline{.916}} & {\textbf{.975}} & {\textbf{.903}} & {\textbf{.016}} & {\textbf{.889}} & {\textbf{.944}} & {\textbf{.870}} & {\textbf{.036}} & {\textbf{.893}} & {\textbf{.953}} & {\textbf{.849}} & {\textbf{.015}} & {\textbf{.903}} & {\textbf{.951}} & {\textbf{.883}} & {\textbf{.023}}\\ 
    \bottomrule
    \end{tabular}
    }
    \label{tab:booktabs1}
\end{table*}

\subsection{Frequency-Based Feature Enhancement Module}
ViT can encode the relationships between all input tokens. However, the feature differences among different tokens in spatial domain are very slight in COD task, making it difficult for the model to discriminate candidate targets.
Thus it is nontrivial to use the more discriminative features learned from the adapter stream to enhance the ViT stream. 
We design the frequency-based feature enhancement (FBFE) module to enhance the features of ViT stream, and take full advantages of both the general knowledge and task-related knowledge. 


As shown in Fig.~\ref{fig:Framework} (c), FGSAttn is first applied to the output of pretrained ViT model $\vF^i_{vit}$, which aims to enhance the target-relevant regions and at the same time suppress background interference with the guidance of frequency domain information. 
Then, we take $\vF^i_{adapter}$ as query to extract the most related 
information from the adjusted general knowledge $\hat{\vF}^i_{vit}$, and obtain the updated adapter feature $\vF^{i+1}_{adapter}$, 
\begin{equation}
\begin{split}
\hat{\vF}^i_{adapter} &= \vF^{i}_{adapter} + Attention(\vF^i_{adapter}, \hat{\vF}^i_{vit}) \\
\vF^{i+1}_{adapter} &= \hat{\vF}^i_{adapter} + FFN(\hat{\vF}^i_{adapter}),
\label{eq6}
\end{split}
\end{equation}
where $Attention( \cdot, \cdot)$ denotes cross-attention mechanism.
$FFN(\cdot)$ denotes the convolutional feed-forward network
to remedy the defect of fixed-size position embeddings~\cite{wang2022pvt}.
After that, the updated adapter feature $\mathbf{F}^{i+1}_{adapter}$ acts as key and value, and $\hat{\mathbf{F}}^i_{vit}$ as query to inject task-related knowledge into ViT feature 
$\mathbf{F}^{i+1}_{vit}$, which will be fed back into the backbone. This process can be expressed as follows:
\begin{equation}
\vF^{i+1}_{vit} = \hat{\vF}^i_{vit} + Attention(\hat{\vF}^i_{vit}, \vF^{i+1}_{adapter}). 
\label{eq7}
\end{equation}
Note that the last FBFE module only outputs the adapter features, which are used for detection.

\subsection{Loss Function}
During training, camouflaged images are fed into both the backbone and adapter simultaneously. 
We only optimize the parameters of the adapter module and detection head, while keeping the parameters of the original pretrained model frozen, so that the power of the ViT foundation model can be efficiently transferred to downstream COD task with little computational cost. Our entire training process is supervised by the combination of weighted binary cross-entropy loss ($L^w_{BCE}$)~\cite{f3net} and weighted intersection-over-union loss ($L^w_{IOU}$)~\cite{f3net}, which can be formulated as $L = L^w_{BCE} + L^w_{IOU}$, forcing the model to pay more attention to hard pixels.

\begin{figure*}[ht!]
    \centering
    \resizebox{\linewidth}{!}{
    \includegraphics{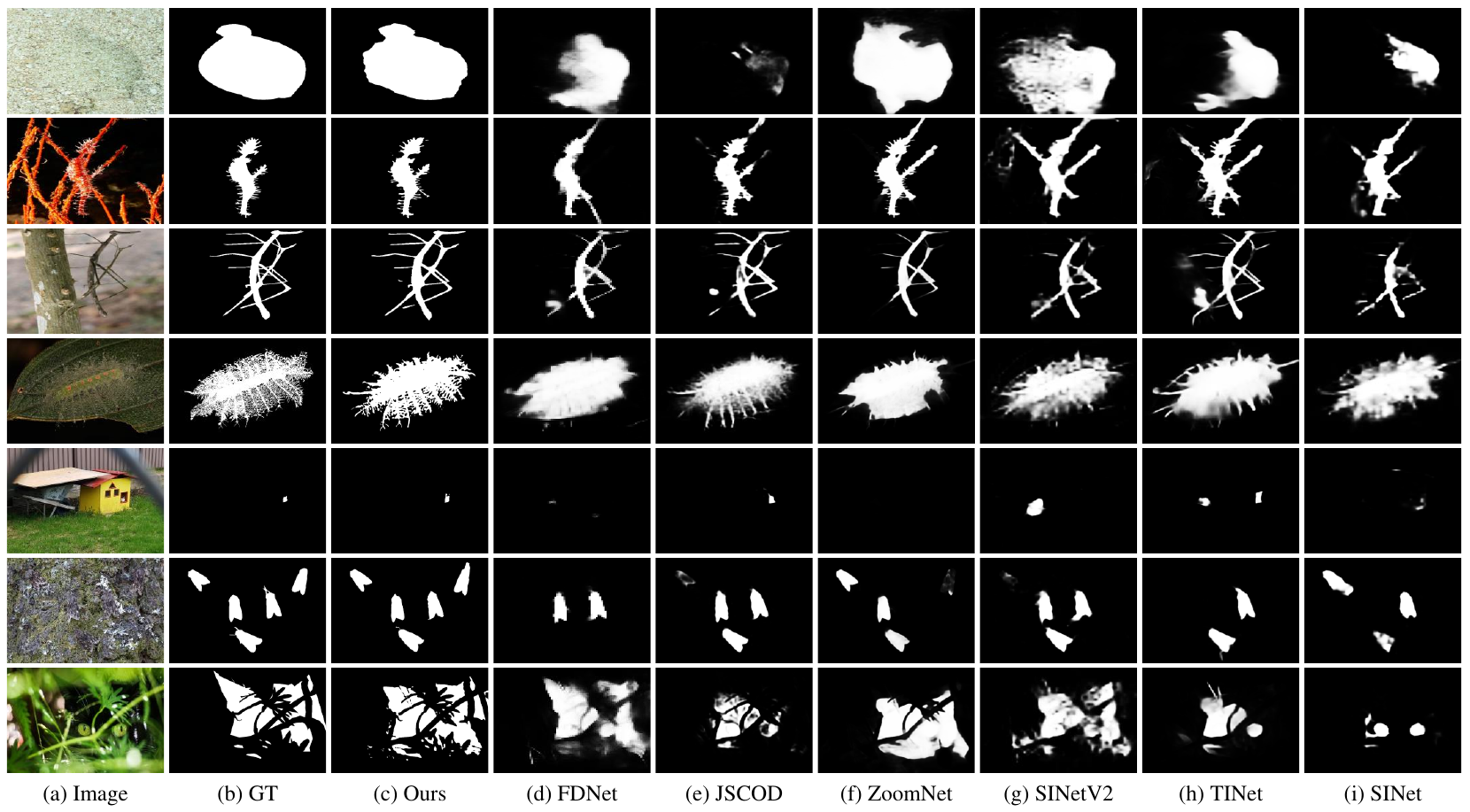}
    }
    \caption{Visual comparison of our method with six representative state-of-the-art methods. Our method is capable of tackling challenging cases (\textit{e.g.}, low contrast, confusing objects, complex and fine structure, indefinable boundary, small object, multiple objects, and occlusion).}
    \label{fig:results}
\end{figure*}
 
\section{Experiments}\label{sec_expr}
\subsection{Experimental Setup}
\paragraph{Datasets}
The experiments are conducted on four benchmark datasets: {CHAMELEON}~\cite{CHAMELEON}, {CAMO}~\cite{CAMO/le2019anabranch},  {COD10K}~\cite{COD10K/SINet}, and {NC4K}~\cite{NC4K/LSR}.
CHAMELEON contains 76 images for test only. While CAMO has 1,000 images for training, and 250 images for testing, consisting of eight categories which fall into both natural and artificial camouflage types.
{COD10K}~\cite{COD10K/SINet} is the largest COD dataset till now, consisting of COD10K-Train (3,040 images) and COD10K-Test (2,026 images). 
{NC4K}~\cite{NC4K/LSR} served as the largest testing dataset which includes 4,121 samples and are typically used to evaluate the generalization ability of models. 
Following experimental protocols in ~\cite{COD10K/SINet}, our method is trained on the training sets of CAMO and COD10K, and the detection performance on the whole CHAMELEON and NC4K datasets, together with the test sets of CAMO and COD10K are reported.

\paragraph{Evaluation metrics}
Four commonly used metrics are adopted for evaluation: Structure measure ($S_{\alpha}$)~\cite{Structure-Measure}, Mean enhanced-alignment measure ($E_{\phi}$)~\cite{E-Measure}, weighted F-measure ($F^w_{\beta}$)~\cite{F-Measure}, and mean absolute error ($M$)~\cite{MAE}.
\paragraph{Training details}
In the training phase, we use Vision Transformer~\cite{ImageWorth2020} as the foundation model and UperNet~\cite{UperNet} as the COD head. 
The Vision Transformer is pretrained with large-scale multi-modal data as in Uni-Perceiver~\cite{Uni-perceiver} and kept frozen once pretrained. 
The parameters of adapter and the COD head are both randomly initialized. We employ an AdamW optimizer with initial learning rate of $6 \times 10^{-5}$ and a weight decay of 0.05. 
They are trained 200 epochs with a batch size of 2. For testing, the images are resized to 512 $\times $512 to input into the model, and the outputs are resized back to the original size.
\paragraph{Competitors}
We compare our method with 26 state-of-the-art COD methods, including:
SINet~\cite{COD10K/SINet}, PraNet~\cite{fan2020pranet}, TINet~\cite{InferringCamouflaged2021}, PFNet~\cite{PFNet}, UGTR~\cite{UncertaintyGuidedTransformer2021}, $\vC^2$FNet~\cite{ContextawareCrosslevel2021}, S-MGL~\cite{MutualGraph2021}, R-MGL~\cite{MutualGraph2021}, LSR~\cite{NC4K/LSR}, JCSOD~\cite{UncertaintyAwareJoint2021}, ERRNet~\cite{ERRNet}, BASNet~\cite{qin2019basnet}, SINetV2~\cite{ConcealedObject2022}, ZoomNet~\cite{ZoomNet}, PENet~\cite{li2023locate/penet}, MFFN~\cite{zheng2023mffn}, FSPNet~\cite{huang2023feature/FSPNet}, HitNet~\cite{HighResolutionIterative2023}, 
DINet~\cite{zhou2024decoupling}, DCT-Net~\cite{zhai2022deep}, UEDG~\cite{lyu2023uedg}, 
FDNet~\cite{DetectingCamouflaged2022}, FBNet~\cite{FrequencyawareCamouflaged2023}, FPNet~\cite{FrequencyPerception2023}, FEDER-MS-4~\cite{he2023camouflaged/FEDER}, SAM-Adapter~\cite{chen2023sam}.
Among these SOTA methods, it is worth noting that  FDNet~\cite{DetectingCamouflaged2022}, FBNet~\cite{FrequencyawareCamouflaged2023}, FPNet~\cite{FrequencyPerception2023}, FEDER-MS-4~\cite{he2023camouflaged/FEDER} all introduce frequency clue from various aspects in their methods. And SAM-Adapter~\cite{chen2023sam} proposed to adapt the Segment Anything foundation model from a spatial perspective without any guidance.
For a fair comparison, all results are either provided by the published paper or reproduced by an open-source model re-trained on the same training set with recommended settings.

\subsection{Comparison With The State-of-the-arts Methods}

\begin{table*}[ht!]
    \caption{Comparison of different parameter-efficient tuning methods with the same pretrained model on four benchmark datasets. The best and second best are \textbf{bolded} and \underline{underlined} for highlighting, respectively.}
    \centering
    \renewcommand\arraystretch{1.3}
    \resizebox{\linewidth}{!}{
    \belowrulesep=3pt
    \aboverulesep=3pt
    \begin{tabular}{c*{4}{c}*{4}{c}*{4}{c}*{4}{c}}
    \toprule
        \multirow{2}*{\textbf{Methods}} & \multicolumn{4}{c}{\textbf{CHAMELEON (76)}} & \multicolumn{4}{c}{\textbf{CAMO-Test (250)}} & \multicolumn{4}{c}{\textbf{COD10K-Test (2,026)}} & \multicolumn{4}{c}{\textbf{NC4K-Test (4,121)}} \\ 
        \cmidrule(lr){2-5}\cmidrule(lr){6-9}\cmidrule(lr){10-13}\cmidrule(lr){14-17}
         ~ &
         $S_{\alpha}\uparrow$ & $E_{\phi}\uparrow$ & $F^w_{\beta}\uparrow$ & 
         $~M~\downarrow$ & 
         $S_{\alpha}\uparrow$ & $E_{\phi}\uparrow$ & $F^w_{\beta}\uparrow$ & 
         $~M~\downarrow$ &
         $S_{\alpha}\uparrow$ & $E_{\phi}\uparrow$ & $F^w_{\beta}\uparrow$ & 
         $~M~\downarrow$ &
         $S_{\alpha}\uparrow$ & $E_{\phi}\uparrow$ & $F^w_{\beta}\uparrow$ & 
         $~M~\downarrow$\\
        \cmidrule(lr){1-17}
         
        Full-Tuning  & \textbf{.923} & \underline{.973} & \textbf{.913} & \underline{.017} & \textbf{.890} & \textbf{.946} & \textbf{.877} & \textbf{.035} & \underline{.891} & .942 & \underline{.845} & \underline{.016} & \underline{.898} & .940 & \underline{.874} & \underline{.024} \\
        Backbone-Frozen  & .870 & .924 & .819 & .028 & .845 & .918 & .815 & .052 & .849 & .916 & .778 & .023 & .870 & .927 & .833 & .032 \\
        \cmidrule(lr){1-17}
        Series-Adapter & .895 & .952 & .866 & .020 & .873 & .936 & .853 & .041 & .877 & .943 & .824 & .018 & .891 & .941 & .863 & .026 \\
        Parallel-Adapter  & .895 & .950 & .867 & .019 & .874 & .936 & .853 & .040 & .874 & .935 & .818 & .018 & .890 & .939 & .863 & .026 \\
        LoRA  & .897 & .949 & .875 & .021 & .879 & .938 & .859 & .040 & .881 & \underline{.944} & .832 & .017 & .894 & \underline{.945} & .871 & .025 \\
        ViT-Adapter  & .909 & .959 & .891 & .018 & .888 & .942 & .868 & .037 & .883 & .943 & .836 & \underline{.016} & .896 & \underline{.945} & \underline{.874} & \underline{.024} \\
        \cmidrule(lr){1-17}
        \textbf{Ours(Frequency-Adapter)} & \underline{.916} & \textbf{.975} & \underline{.903} & \textbf{.016} & \underline{.889} & \underline{.944} & \underline{.870} & \underline{.036} & \textbf{.893} & \textbf{.953} & \textbf{.849} & \textbf{.015} & \textbf{.903} & \textbf{.951} & \textbf{.883} & \textbf{.023} \\

    \bottomrule
    \end{tabular}
    }
    \label{tab:booktabs2}
\end{table*}

\begin{table*}[ht!]
    \caption{Ablation studies of the core operation in FGSAttn. best results are marked in \textbf{bold} fonts.}
    \centering
    \renewcommand\arraystretch{1.3}
    \resizebox{\linewidth}{!}{
    \belowrulesep=3pt
    \aboverulesep=3pt
    \begin{tabular}{c*{4}{c}*{4}{c}*{4}{c}*{4}{c}}
    \toprule
        \multirow{2}*{\textbf{Core operation}} & \multicolumn{4}{c}{\textbf{CHAMELEON (76)}} & \multicolumn{4}{c}{\textbf{CAMO-Test (250)}} & \multicolumn{4}{c}{\textbf{COD10K-Test (2,026)}} & \multicolumn{4}{c}{\textbf{NC4K-Test (4,121)}} \\ 
        
        \cmidrule(lr){2-5}\cmidrule(lr){6-9}\cmidrule(lr){10-13}\cmidrule(lr){14-17}
         ~ & 
         $S_{\alpha}\uparrow$ & $E_{\phi}\uparrow$ & $F^w_{\beta}\uparrow$ & 
         $~M~\downarrow$ & 
         $S_{\alpha}\uparrow$ & $E_{\phi}\uparrow$ & $F^w_{\beta}\uparrow$ & 
         $~M~\downarrow$ &
         $S_{\alpha}\uparrow$ & $E_{\phi}\uparrow$ & $F^w_{\beta}\uparrow$ & 
         $~M~\downarrow$ &
         $S_{\alpha}\uparrow$ & $E_{\phi}\uparrow$ & $F^w_{\beta}\uparrow$ & 
         $~M~\downarrow$ \\

        \cmidrule{1-17}
        Regular convolution	& .910 & .964 & .893 & .018 & .874 & .938 & .862 & .038 & .888 & .946 & .839 & .016 & .899 & .948 & .875 & .025 \\
        Deformable convolution & .912 & .964 & .895 & .017 & .873 & .939 & .862 & .037 & .889 & .945 & .840 & .016 & .897 & .947 & .878 & .025 \\
        \cmidrule{1-17}
        {\textbf{Ours(Frequency-Based)}}& {\textbf{.916}} & {\textbf{.975}} & {\textbf{.903}} & {\textbf{.016}} & {\textbf{.889}} & {\textbf{.944}} & {\textbf{.870}} & {\textbf{.036}} & {\textbf{.893}} & {\textbf{.953}} & {\textbf{.849}} & {\textbf{.015}} & {\textbf{.903}} & {\textbf{.951}} & {\textbf{.883}} & {\textbf{.023}}\\ 

    \bottomrule
    \end{tabular}
    }
    \label{tab:booktabs3}
\end{table*}

\paragraph{Quantitative evaluation}

Table~\ref{tab:booktabs1} reports the detailed comparision results of our FGSA-Net against other 26 state-of-the-art methods on four benchmark datasets. 
It can be seen that our proposed method outperforms all the comparison SOTA methods with a large margin on all the benchmark datasets.
For example, our method achieves 0.893 $S_{\alpha}$ and 0.849 $F^w_{\beta}$ on COD10K dataset, greatly outperforms the second best SAM-adapter method. 
And on NC4K dataset, our method sets a remarkable record to increase $S_{\alpha}$ by 2.50\%, $E_{\phi}$ by 2.48\%, $F^w_{\beta}$ by 6.51\% and lowers the MAE error by 34.3\%, compared with the second best UEDG method.
It is worth noting that, our method greatly outperforms SAM-adapter method which is also based on a vision foundation model (SAM) and tuned with a spatial adapter.
As SAM-Adapter mainly learns task specific knowledge and injects novel knowledge of downstream task into the model through the adapter from the perspective of spatial domain alone.
Due to the high similarity between the camouflaged object and the surrounding environment, spatial adaptation with the guidance from spatial domain directly is easily confused and can not effectively extract subtle features. 
By contrast, introducing task specific knowledge into the model under the guidance of frequency domain can enable the network to pay more attention to concealed targets.
Furthermore, our method also outperforms all the existing frequency-based methods, namely FDNet, FBNet, FPNet and FEDER-MS-4, on all four standard metrics.
Compared to other frequency-based methods, our advantage lies in fully utilizing the general knowledge of vision foundation model, proving that the designed adapter can effectively transfer vision foundation model into downstream tasks such as COD.

\begin{table*}[ht!]
    \caption{Ablation studies of FGSAttn in different modules. Best results are marked in \textbf{bold}.}
    \centering
    \renewcommand\arraystretch{1.3}
    \resizebox{\linewidth}{!}{
    \belowrulesep=3pt
    \aboverulesep=3pt
    \begin{tabular}{c*{2}{c}*{4}{c}*{4}{c}*{4}{c}*{4}{c}}
    \toprule
        \multirow{2}*{\textbf{Settings}} & \multicolumn{2}{c}{\textbf{Components}} & \multicolumn{4}{c}{\textbf{CHAMELEON (76)}} & \multicolumn{4}{c}{\textbf{CAMO-Test (250)}} & \multicolumn{4}{c}{\textbf{COD10K-Test (2,026)}} & \multicolumn{4}{c}{\textbf{NC4K-Test (4,121)}} \\ 
        
        \cmidrule(lr){2-3}\cmidrule(lr){4-7}\cmidrule(lr){8-11}\cmidrule(lr){12-15}\cmidrule(lr){16-19}
         ~ & \textbf{FBNM} & \textbf{FBFE} &
         $S_{\alpha}\uparrow$ & $E_{\phi}\uparrow$ & $F^w_{\beta}\uparrow$ & 
         $~M~\downarrow$ & 
         $S_{\alpha}\uparrow$ & $E_{\phi}\uparrow$ & $F^w_{\beta}\uparrow$ & 
         $~M~\downarrow$ &
         $S_{\alpha}\uparrow$ & $E_{\phi}\uparrow$ & $F^w_{\beta}\uparrow$ & 
         $~M~\downarrow$ &
         $S_{\alpha}\uparrow$ & $E_{\phi}\uparrow$ & $F^w_{\beta}\uparrow$ & 
         $~M~\downarrow$\\
         
         \cmidrule(lr){1-19} 
        Baseline & ~ & ~ & 
         .909 & .959 & .891 & .018 & .878 & .935 & .868 & .037 & .883 & .943 & .836 & .016 & .896 & .945 & .874 & .024 \\
        Variant 1 & ~ & \checkmark & .910 & .969 & .892 & .019 & .884 & .942 & .866 & .037 & .887 & .949 & .839 & .016 & .898 & .946 & .873 & .024 \\
        Variant 2 & \checkmark & ~ & .910 & .969 & .894 & .018 & .887 & .943 & .866 & .037 & .890 & .951 & .845 & .016 & .899 & .947 & .875 & .025 \\
       
        \cmidrule(lr){1-19}
        \textbf{Ours(FGSA-Net)} & \checkmark & \checkmark & {\textbf{.916}} & {\textbf{.975}} & {\textbf{.903}} & {\textbf{.016}} & {\textbf{.889}} & {\textbf{.944}} & {\textbf{.870}} & {\textbf{.036}} & {\textbf{.893}} & {\textbf{.953}} & {\textbf{.849}} & {\textbf{.015}} & {\textbf{.903}} & {\textbf{.951}} & {\textbf{.883}} & {\textbf{.023}}\\ 
    \bottomrule
    \end{tabular}
    }
    \label{tab:booktabs4}
\end{table*}

\paragraph{Qualitative evaluation}

In Fig.~\ref{fig:results}, we show the qualitative comparison of our method with several representative SOTA methods on some challenging situations.
Benefiting from the discriminative frequency information, our FGSA-Net achieves more competitive visual performance mainly in the following aspects: More accurate localization and complete prediction of targets in low contrast scenes (Row 1), stronger interference suppression when there are confusing objects in the surrounding environment (Row 2) and more precise recognition of complex and fine structure, such as slender details of the object (Row 3). 
Moreover, our method is also effective in detecting other challenging situations such as indefinable boundary, small object, multiple objects and occlusion (Row 4 to Row 7). The impressive prediction results further verified the effectiveness of the frequency-guided spatial adaptation network.

\subsection{Further Analysis}

\paragraph{Effectiveness of frequency-guided adapter}
To show the effectiveness of our FGSA-Net, we implement other four types of adapters, namely Series-Adapter, Parallel-Adapter, LoRA and ViT-Adapter, while keeping the same pretrained foundation model and pretrained weights with our method. The results in Table~\ref{tab:booktabs2} show that the proposed FGSA-Net can greatly outperform all the spatial adaptation variants on four benchmark datasets, indicating that through frequency-guided spatial adaptation on COD task, the general knowledge can be better transferred to deal with the COD problem.

\paragraph{Core operation in FGSAttn}
To show the advantage of our frequency-based attention over other spatial attention, we compare our method with two variants, which are replacing core operation in FGSAttn with a regular convolution module and a deformable convolution module, respectively. The results are shown in Table~\ref{tab:booktabs3}. It can be seen that our method performs better on all datasets compared with other variants. We analyze the reason is that the brightness difference between camouflaged object and surrounding environment is very small in the spatial domain, and attention maps generated by using convolution operations based on spatial domain may be confused, making it difficult for the model to effectively focus on the camouflaged object and detailed clues. 

\paragraph{Effect of FGSAttn on FBNM and FBFE}
To show how much our proposed FGSAttn takes effect on FBNM and FBFE module, we evaluate our method while removing FGSAttn in FBNM and FBFE respectively. 
As can be seen from Table~\ref{tab:booktabs4} that removing FGSAttn in either module would get a significant degradation of our FGSA-Net.

\begin{table*}[ht!]
    \caption{Effectiveness of various input sizes. Best results are marked in \textbf{bold}.}
    \centering
    \renewcommand\arraystretch{1.3}
    \resizebox{\linewidth}{!}{
    \belowrulesep=3pt
    \aboverulesep=3pt
    \begin{tabular}{c*{4}{c}*{4}{c}*{4}{c}*{4}{c}}
    \toprule
        \multirow{2}*{\textbf{Input size}} & \multicolumn{4}{c}{\textbf{CHAMELEON (76)}} & \multicolumn{4}{c}{\textbf{CAMO-Test (250)}} & \multicolumn{4}{c}{\textbf{COD10K-Test (2,026)}} & \multicolumn{4}{c}{\textbf{NC4K-Test (4,121)}} \\ 
        
        \cmidrule(lr){2-5}\cmidrule(lr){6-9}\cmidrule(lr){10-13}\cmidrule(lr){14-17}
         ~ &
         $S_{\alpha}\uparrow$ & $E_{\phi}\uparrow$ & $F^w_{\beta}\uparrow$ & 
         $~M~\downarrow$ & 
         $S_{\alpha}\uparrow$ & $E_{\phi}\uparrow$ & $F^w_{\beta}\uparrow$ & 
         $~M~\downarrow$ &
         $S_{\alpha}\uparrow$ & $E_{\phi}\uparrow$ & $F^w_{\beta}\uparrow$ & 
         $~M~\downarrow$ &
         $S_{\alpha}\uparrow$ & $E_{\phi}\uparrow$ & $F^w_{\beta}\uparrow$ & 
         $~M~\downarrow$ \\

        \cmidrule{1-17}
        352$\times$352 & .901 & .963 & .884 & .019 & .880 & .939 & .857 & .040 & .873 & .939 & .813 & .019 & .893 & .944 & .865 & .025 \\
        384$\times$384 & .901 & .969 & .888 & .020 & .881 & .938 & .858 & .038 & .879 & .946 & .824 & .018 & .897 & .947 & .870 & .025 \\
        416$\times$416 & .908 & .971 & .894 & .018 & .882 & .940 & .861 & .038 & .883 & .947 & .831 & .017 & .898 & .948 & .873 & .024 \\
        \cmidrule{1-17}
        512$\times$512 & {\textbf{.916}} & {\textbf{.975}} & {\textbf{.903}} & {\textbf{.016}} & {\textbf{.889}} & {\textbf{.944}} & {\textbf{.870}} & {\textbf{.036}} & {\textbf{.893}} & {\textbf{.953}} & {\textbf{.849}} & {\textbf{.015}} & {\textbf{.903}} & {\textbf{.951}} & {\textbf{.883}} & {\textbf{.023}}\\ 

    \bottomrule
    \end{tabular}
    }
    \label{tab:booktabs5}
\end{table*}

\begin{table*}[ht!]
    \caption{Effectiveness of different pretrained weights. Best results are marked in \textbf{bold}.}
    \centering
    \renewcommand\arraystretch{1.3}
    \resizebox{\linewidth}{!}{
    \belowrulesep=3pt
    \aboverulesep=3pt
    \begin{tabular}{c*{4}{c}*{4}{c}*{4}{c}*{4}{c}}
    \toprule
        \multirow{2}*{\textbf{Backbone}} & \multicolumn{4}{c}{\textbf{CHAMELEON (76)}} & \multicolumn{4}{c}{\textbf{CAMO-Test (250)}} & \multicolumn{4}{c}{\textbf{COD10K-Test (2,026)}} & \multicolumn{4}{c}{\textbf{NC4K-Test (4,121)}} \\ 
        
        \cmidrule(lr){2-5}\cmidrule(lr){6-9}\cmidrule(lr){10-13}\cmidrule(lr){14-17}
         ~ &
         $S_{\alpha}\uparrow$ & $E_{\phi}\uparrow$ & $F^w_{\beta}\uparrow$ & 
         $~M~\downarrow$ & 
         $S_{\alpha}\uparrow$ & $E_{\phi}\uparrow$ & $F^w_{\beta}\uparrow$ & 
         $~M~\downarrow$ &
         $S_{\alpha}\uparrow$ & $E_{\phi}\uparrow$ & $F^w_{\beta}\uparrow$ & 
         $~M~\downarrow$ &
         $S_{\alpha}\uparrow$ & $E_{\phi}\uparrow$ & $F^w_{\beta}\uparrow$ & 
         $~M~\downarrow$ \\

        \cmidrule{1-17}
        AugReg & .910 & .964 & .895 & .018 & .875 & .939 & .853 & .040 & .883 & .934 & .833 & .017 & .891 & .933 & .863 & .024 \\
        BEiT & .911 & .966 & .898 & .017 & .871 & .930 & .847 & .046 & .881 & .932 & .825 & .019 & .889 & .936 & .862 & .026 \\
        SAM & .915 & .969 & .901 & .017 & .873 & .938 & .853 & .043 & .889 & .936 & .839 & .018 & .893 & .934 & .865 & .027 \\
        \cmidrule{1-17}
        Uni-Perceiver & {\textbf{.916}} & {\textbf{.975}} & {\textbf{.903}} & {\textbf{.016}} & {\textbf{.889}} & {\textbf{.944}} & {\textbf{.870}} & {\textbf{.036}} & {\textbf{.893}} & {\textbf{.953}} & {\textbf{.849}} & {\textbf{.015}} & {\textbf{.903}} & {\textbf{.951}} & {\textbf{.883}} & {\textbf{.023}}\\ 

    \bottomrule
    \end{tabular}
    }
    
    \label{tab:booktabs6}
\end{table*}

\begin{table*}[ht!]
    \caption{Effectiveness of different K and M. Best results are marked in \textbf{bold}.}
    \centering
    \renewcommand\arraystretch{1.3}
    \resizebox{\linewidth}{!}{
    \belowrulesep=3pt
    \aboverulesep=3pt
    \begin{tabular}{cc*{4}{c}*{4}{c}*{4}{c}*{4}{c}}
    \toprule
        \multirow{2}*{\textbf{K}} & \multirow{2}*{\textbf{M}} & \multicolumn{4}{c}{\textbf{CHAMELEON(76)}} & \multicolumn{4}{c}{\textbf{CAMO-Test(250)}} & \multicolumn{4}{c}{\textbf{COD10K-Test(2,026)}} & \multicolumn{4}{c}{\textbf{NC4K-Test(4,121)}} \\ 
        \cmidrule(lr){3-6}\cmidrule(lr){7-10}\cmidrule(lr){11-14}\cmidrule(lr){15-18}
         ~ & ~ &
         $S_{\alpha}\uparrow$ & $E_{\phi}\uparrow$ & $F^w_{\beta}\uparrow$ & 
         $~M~\downarrow$ & 
         $S_{\alpha}\uparrow$ & $E_{\phi}\uparrow$ & $F^w_{\beta}\uparrow$ & 
         $~M~\downarrow$ &
         $S_{\alpha}\uparrow$ & $E_{\phi}\uparrow$ & $F^w_{\beta}\uparrow$ & 
         $~M~\downarrow$ &
         $S_{\alpha}\uparrow$ & $E_{\phi}\uparrow$ & $F^w_{\beta}\uparrow$ & 
         $~M~\downarrow$\\
        \cmidrule(lr){1-18}
        3 & 8 & .910 & .971 & .900 & .018 & .884 & .939 & .866 & .038 & .884 & .949 & .840 & .019 & .899 & .941 & .872 & .025 \\
        4 & 6 & .915 & .973 & \textbf{.903} & .017 & \textbf{.890} & .942 & .868 & \textbf{.036} & .891 & .950 & .846 & .016 & .902 & \textbf{.952} & .882 & \textbf{.023} \\ 
        6 & 4 & \textbf{.916} & \textbf{.975} & \textbf{.903} & \textbf{.016} & .889 & \textbf{.944} & \textbf{.870} & \textbf{.036} & \textbf{.893} & \textbf{.953} & \textbf{.849} & \textbf{.015} & \textbf{.903} & .951 & \textbf{.883} & \textbf{.023} \\
        8 & 3 & .906 & .968 & .898 & .019 & .881 & .936 & .865 & .038 & .885 & .947 & .842 & .018 & .899 & .942 & .875 & .026 \\
    \bottomrule
    \end{tabular}
    }
    \label{tab:K}
\end{table*}
\begin{table*}[ht!]
   \caption{Effectiveness of different width in the amplitude decomposition. Best results are marked in \textbf{bold}.}
    \centering
    \renewcommand\arraystretch{1.3}
    \resizebox{\linewidth}{!}{
    \belowrulesep=3pt
    \aboverulesep=3pt
    \begin{tabular}{c*{4}{c}*{4}{c}*{4}{c}*{4}{c}}
    \toprule
        \multirow{2}*{\textbf{d}} & \multicolumn{4}{c}{\textbf{CHAMELEON(76)}} & \multicolumn{4}{c}{\textbf{CAMO-Test(250)}} & \multicolumn{4}{c}{\textbf{COD10K-Test(2,026)}} & \multicolumn{4}{c}{\textbf{NC4K-Test(4,121)}} \\ 
        \cmidrule(lr){2-5}\cmidrule(lr){6-9}\cmidrule(lr){10-13}\cmidrule(lr){14-17}
         ~ &
         $S_{\alpha}\uparrow$ & $E_{\phi}\uparrow$ & $F^w_{\beta}\uparrow$ & 
         $~M~\downarrow$ & 
         $S_{\alpha}\uparrow$ & $E_{\phi}\uparrow$ & $F^w_{\beta}\uparrow$ & 
         $~M~\downarrow$ &
         $S_{\alpha}\uparrow$ & $E_{\phi}\uparrow$ & $F^w_{\beta}\uparrow$ & 
         $~M~\downarrow$ &
         $S_{\alpha}\uparrow$ & $E_{\phi}\uparrow$ & $F^w_{\beta}\uparrow$ & 
         $~M~\downarrow$\\
        \cmidrule(lr){1-17}
        1 & \textbf{.916} & \textbf{.975} & \textbf{.903} & \textbf{.016} & \textbf{.889} & \textbf{.944} & \textbf{.870} & .036 & .893 & \textbf{.953} & \textbf{.849} & \textbf{.015} & \textbf{.903} & \textbf{.951} & \textbf{.883} & \textbf{.023} \\
        2 & .915 & .971 & .901 & .017 & .887 & .942 & .869 & \textbf{.035} & \textbf{.895} & .952 & .847 & .016 & .902 & .949 & .882 & \textbf{.023} \\ 
        4 & .912 & .970 & .899 & .018 & .888 & .943 & .868 & .036 & .892 & .952 & .847 & .016 & .902 & .950 & .881 & .024 \\
        8 & .910 & .969 & .899 & .018 & .882 & .942 & .865 & .037 & .890 & .949 & .846 & .017 & .901 & .948 & .879 & .024 \\
    \bottomrule
    \end{tabular}
    }
    \label{tab:d}
\end{table*}

\paragraph{Feature map visualization after adaptation}
Fig.~\ref{fig:fig1} illustrates some representative cases of the obtained feature maps after adapter tuning on COD task.  It is noticeable that series-adapter, parallel-adapter and LoRA can only focus on the boundaries of the target roughly, while {ViT-adapter} highlights the target but also generates more background noise, which is detrimental to the model's localization and recognition capabilities. 
Only our proposed frequency-guided spatial adaptation method clearly tunes the pretrained model to focus more on the concealed foreground objects compared with other four spatial adaptation counterparts. 

\paragraph{Various input image sizes}
To explore the impact of various input image sizes on model performance, we present the results of our model at different input sizes, including 352 $\times $ 352, 384 $\times $ 384, 416 $\times $ 416 and 512 $\times $ 512, which are illustrated in Table~\ref{tab:booktabs5}. As can be seen, the performance of the model gradually improves with the increase of input image resolution, and our model performs the best at a setting of 512 $\times $ 512. 
It is worth noting that when decreasing input size into 416 $\times $ 416, 384 $\times $ 384, 352 $\times $ 352, our method also achieve SOTA results on three datasets(CAMO, COD10K and NC4K) and competitive performance on CHAMELEON dataset, even though other models use larger input sizes, such as FPNet use 512 $\times $ 512, HitNet use 704 $\times $ 704, and ZoomNet use multiple inputs (maximum 576 $\times $ 576). 

\paragraph{Different pretrained weights}
In order to explore the impact of different pretrained weights, we experiment with ViT as the backbone and initialize it with different pretrained weights, including AugReg~\cite{AugReg} which trained on ImageNet-22K, BEiT~\cite{Beit} which trained on ImageNet-1K, SAM~\cite{kirillov2023segany} which trained on 11 million images and 1.1 billion masks, and Uni-Perceiver~\cite{Uni-perceiver} which is trained with large scale multi-modal data.
As summarized in Table~\ref{tab:booktabs6}, we find that using various pretrained weights both achieve competitive performance, which verifies the effectiveness of our designed adapter for different pretrained backbone. Among them, the backbone pretrained with multi-modal data show the best performance. It is worth noting that our method significantly outperforms the SAM-Adapter method by utilizing the SAM initialized backbone, demonstrating the superiority of our proposed FGSA-Net.

\begin{table*}[ht!]
    \caption{Quantitative comparison of our FGSA-Net and 15 SOTA methods for SOD on four benchmark datasets. $\uparrow$ / $\downarrow$ indicates that larger/smaller is better. The best and second best are \textbf{bolded} and \underline{underlined} for highlighting, respectively.}
    \centering
    \renewcommand\arraystretch{1.3}
    \resizebox{\linewidth}{!}{
    \belowrulesep=3pt
    \aboverulesep=3pt
    \begin{tabular}{c*{4}{c}*{4}{c}*{4}{c}*{4}{c}}
    \toprule
        \multirow{2}*{\textbf{Methods}} & \multicolumn{4}{c}{\textbf{ECSSD}} & \multicolumn{4}{c}{\textbf{DUTS-TE}} & \multicolumn{4}{c}{\textbf{HKU-IS}} & \multicolumn{4}{c}{\textbf{DUT-OMRON}} \\ 
        \cmidrule(lr){2-5}\cmidrule(lr){6-9}\cmidrule(lr){10-13}\cmidrule(lr){14-17}
        
         ~ & 
         $S_{\alpha}\uparrow$ & $E_{\phi}\uparrow$ & $F^w_{\beta}\uparrow$ & 
         $~M~\downarrow$ & 
         $S_{\alpha}\uparrow$ & $E_{\phi}\uparrow$ & $F^w_{\beta}\uparrow$ & 
         $~M~\downarrow$ &
         $S_{\alpha}\uparrow$ & $E_{\phi}\uparrow$ & $F^w_{\beta}\uparrow$ & 
         $~M~\downarrow$ &
         $S_{\alpha}\uparrow$ & $E_{\phi}\uparrow$ & $F^w_{\beta}\uparrow$ & 
         $~M~\downarrow$ \\
         \cmidrule(lr){1-17} 
        
        BMPM & .911 & .914 & .871 & .045 & .862 & .860 & .761 & .049 & .907 & .937 & .859 & .039 & .809 & .837 & .681 & .064 \\
        RAS & .893 & .914 & .857 & .056 & .839 & .861 & .740 & .059 & .887 & .929 & .843 & .045 & .814 & .846 & .695 & .062 \\
        PiCA-R & .917 & .913 & .867 & .046 & .869 & .862 & .754 & .043 & .904 & .936 & .840 & .043 & .832 & .841 & .695 & .065 \\
        DGRL & .906 & .917 & .903 & .043 & .846 & .863 & .764 & .051 & .896 & .941 & .881 & .037 & .810 & .843 & .709 & .063 \\
        CPD-R & .918 & .925 & .898 & .037 & .869 & .886 & .795 & .043 & .905 & .944 & .875 & .034 & .825 & .866 & .719 & .056 \\
        PoolNet & .926 & .925 & .904 & .035 & .886 & .896 & .817 & .037 & .919 & .953 & .888 & .030 & .831 & .868 & .725 & .054 \\
        SIBA & .924 & .928 & .908 & .035 & .879 & .892 & .811 & .040 & .913 & .950 & .886 & .032 & .832 & .860 & .736 & .059 \\
        EGNet & .925 & .927 & .903 & .037 & .887 & .891 & .815 & .039 & .918 & .950 & .887 & .031 & .841 & .867 & .738 & .053 \\
        F3Net & .924 & .925 & .912 & .034 & .888 & .902 & .835 & .035 & .917 & .953 & .900 & .028 & .838 & .870 & .747 & .053 \\
        ICON & .931 & .924 & .920 & .031 & .892 & .900 & .839 & .037 & .925 & .956 & .908 & .027 & .845 & .866 & .762 & .058 \\

        TSNet & .936 & .917 & .915 & .036 & .883 & .854 & .804 & .038 & .921 & .912 & .902 & .031 & .858 & .809 & .761 & .044 \\
        PRNet & .917 & .946 & .895 & .039 & .879 & .910 & .811 & .039 & .913 & .956 & .885 & .032 & .829 & .866 & .723 & .056 \\
        
        TINet & .926 & .953 & .914 & .033 & .891 & .925 & .842 & .035 & .922 & .960 & .906 & .027 & .842 & .876 & .754 & .051 \\
        JCSOD & .933 & .960 & .935 & .030 & .899 & .937 & .866 & .032 & \underline{.931} & .867 & \underline{.924} & .026 & .850 & .884 & .782 & .051 \\
        BIPGNet & \underline{.938} & \underline{.962} & \underline{.938} & \underline{.025} & \underline{.905} & \underline{.940} & \underline{.877} & \underline{.029} & .924 & \underline{.964} & .922 & \underline{.023} & \underline{.854} & \underline{.886} & \underline{.787} & \underline{.047} \\
        
        \cmidrule(lr){1-17}
        
        \textbf{Ours(FGSA-Net)} & \textbf{.949} & \textbf{.981} & \textbf{.952} & \textbf{.019} & \textbf{.926} & \textbf{.956} & \textbf{.899} & \textbf{.024} & \textbf{.939} & \textbf{.981} & \textbf{.939} & \textbf{.016} & \textbf{.861} & \textbf{.892} & \textbf{.796} & \textbf{.045} \\  
    \bottomrule
    \end{tabular}
    }
    \label{tab:booktabs7}
\end{table*}

\paragraph{Different K and M}
{For the pretrained ViT model with $L$=24, we study the effect of $K$ and $M$, and the results are shown in Table~\ref{tab:K}. It can be seen that the model achieves optimal performance on most datasets and metrics when $K$=6 and $M$=4, and dividing into more groups cannot bring significant gains. Therefore, we empirically set $K$=6 and $M$=4.}

\paragraph{Different d in the amplitude decomposition}
{The specific value for width $d$ represents the range of frequency components contained in each channel. To explore the influence of $d$ in the amplitude decomposition, we present the results of our model at different widths, as shown in table~\ref{tab:d}. It can be observed that the performance is the best when $d$=1, achieving the highest score on multiple metrics, and the performance gradually decreases with the increase of $d$. We speculate that finer decomposition of amplitude is beneficial for the model to better adaptively adjust various attributes of objects, such as the approximate shape or edge details, etc.}

\paragraph{Generalization performance on salient object detection (SOD) task}
To validate the generalization of our method on SOD task, we train our FGSA-Net on the DUTS-TR~\cite{DUTS} dataset, and directly evaluate on other four testing datasets, including ECSSD~\cite{ECSSD}, DUTS-TE~\cite{DUTS}, HKU-IS~\cite{HKU-IS} and DUT-OMRON~\cite{OMRON}.
We compare our method with 15 representative methods, including BMPM~\cite{BMPM}, RAS~\cite{RAS}, PiCA-R~\cite{picanet}, DGRL~\cite{DGRL}, CPD-R~\cite{CPD-R}, PoolNet~\cite{PoolNet}, SIBA~\cite{SIBA}, EGNet~\cite{EGNet}, F3Net~\cite{f3net}, ICON~\cite{ICON}, TSNet~\cite{TSNet}, PRNeT~\cite{PRNet}, TINet~\cite{TINet} and JCSOD~\cite{JCSOD}, BIPGNet~\cite{BIPGNet}.
Table~\ref{tab:booktabs7} reports the quantitative results on four SOD benchmark datasets. It can be seen that our model performs favorably against the existing methods in terms of nearly all evaluation metrics. 
For example, compared with the second-best model BIPGNet on ECSSD dataset, our model increases $S_{\alpha}$, $E_{\phi}$, and $F^w_{\beta}$ by 1.17\%, 1.98\%, and 1.50\% respectively, and lowers the MAE error by 24\%.
This demonstrates the strong capability and effectiveness of our network to deal with other binary segmentation task.

\begin{table*}[ht!]
    \caption{Parameter comparison of our FGSA-Net with 11 representative state-of-the-art methods.}
    \centering
    \renewcommand\arraystretch{1.3}
    \resizebox{\linewidth}{!}{
    \belowrulesep=3pt
    \aboverulesep=3pt
    \begin{tabular}{ccccccccccccc}
    \toprule
        ~ & \textbf{Ours} & SINetV2 & ZoomNet & PFNet & SINet & LSR & S-MGL & R-MGL & ERRNet & BASNet & JSCOD & SAM-Adapter \\
        \cmidrule(lr){1-13}
        \#Param(M) & \textbf{59.90} & 26.98 & 32.38 & 46.50 & 48.95 & 57.90 & 63.60 & 67.64 & 69.76 & 87.06 & 121.63 & 206.3\\
    \bottomrule
    \end{tabular}
    }
    \label{tab:booktabs8}
\end{table*}

\paragraph{Parameter comparison}
In table~\ref{tab:booktabs8}, we compare the number of tunable parameters of our method and some representative SOTA methods, including SINetV2, ZoomNet, PFNet, SINet, LSR, S-MGL, R-MGL, ERRNet, BASNet, JSCOD and SAM-Adapter. {For fair comparison, all parameter results are either provided in the published paper or calculated based on the implementation details in the paper and open-source model code.} These statistics highlight that our proposed FGSA-Net is lightweight, requiring less or comparable parameters to achieve promising performance. 

\section{Conclusion}\label{sec_conclusion}
In this paper, we propose a frequency-guided spatial adaptation network for COD. 
Specifically, a frequency-guided spatial attention module is devised to adapt the pretrained foundation model from spatial domain to focus more on the camouflaged regions, while guided by the frequency components dynamically adjusted in the frequency domain.
Based on the attention module, the FBNM and FBFE module are further proposed to extract and fuse multi-scale features which contain both the general knowledge of the pretrained model and specialized knowledge learned from the downstream COD dataset.
Extensive experiments verify that our proposed method outperforms the baseline counterparts with large margins and achieves state-of-the-art performances on four benchmark datasets.


\bibliographystyle{IEEEtran}
\bibliography{egbib}

\begin{IEEEbiography}[{\includegraphics[width=1in,height=1.25in,clip,keepaspectratio]{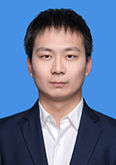}}]{Shizhou Zhang}
received a B.E. and Ph.D. degree from Xi'an Jiaotong University, Xi'an, China, in 2010 and 2017, respectively. Currently, he is with Northwestern Polytechnical University as an Associate Professor (Tenured). His research interests include content-based image analysis, pattern recognition and machine learning, specifically in the areas of deep learning-based vision tasks such as image classification, object detection, re-identification.
\end{IEEEbiography}

\vspace{-25pt}

\begin{IEEEbiography}[{\includegraphics[width=1in,height=1.25in,clip,keepaspectratio]{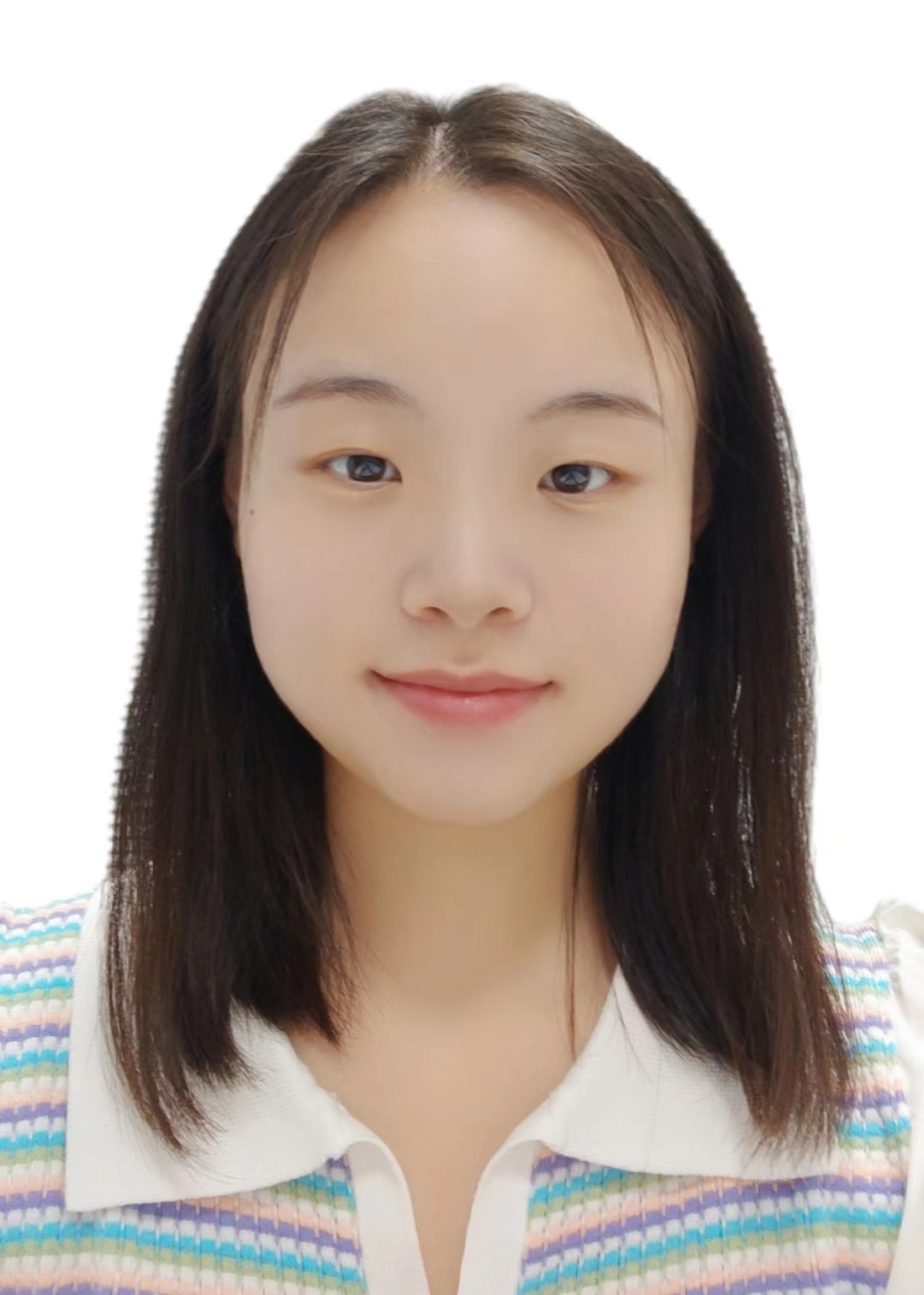}}]{Dexuan Kong}
received a B.E. degree from the School of Computer Science, Hebei Normal University, in 2022. She is currently working toward an M.E. degree with the School of Computer Science, Northwestern Polytechnical University. Her research interests include camouflaged object detection, and remote sensing image processing.
\end{IEEEbiography}

\vspace{-25pt}

\begin{IEEEbiography}[{\includegraphics[width=1in,height=1.25in,clip,keepaspectratio]{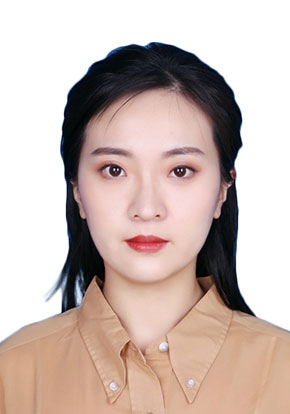}}]{Yinghui Xing}
received a B.S. and Ph.D. degrees from the School of Artificial Intelligence, Xidian University, Xi'an, China, in 2014 and 2020, respectively. She is now an Associate Professor with the School of Computer Science, Northwestern Polytechnical University, Xi'an. Her research interests include remote sensing image processing, image fusion, and image super resolution.

\end{IEEEbiography}

\vspace{-25pt}
\begin{IEEEbiography}[{\includegraphics[width=1in,height=1.25in,clip,keepaspectratio]{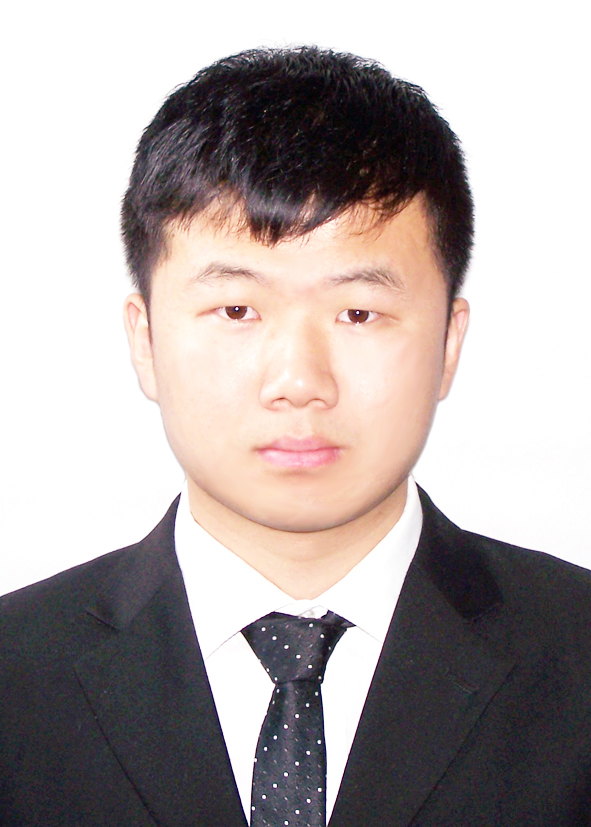}}]
{Yue Lu} received the B.E. degree in automation from Beijing Information Science and Technology University, Beijing, China, in 2020, and the M.E. degree in computer science from Northwestern Polytechnical University,  Xi'an, China, in 2023. He is currently pursuing the Ph.D degree in computer science with the National Engineering Laboratory for Integrated Aero-Space-Ground-Ocean Big Data Application Technology, School of Computer Science, Northwestern Polytechnical University, Xi'an, China. His current research interests include computer vision and continual learning.

\end{IEEEbiography}

\vspace{-25pt}

\begin{IEEEbiography}[{\includegraphics[width=1in,height=1.25in,clip,keepaspectratio]{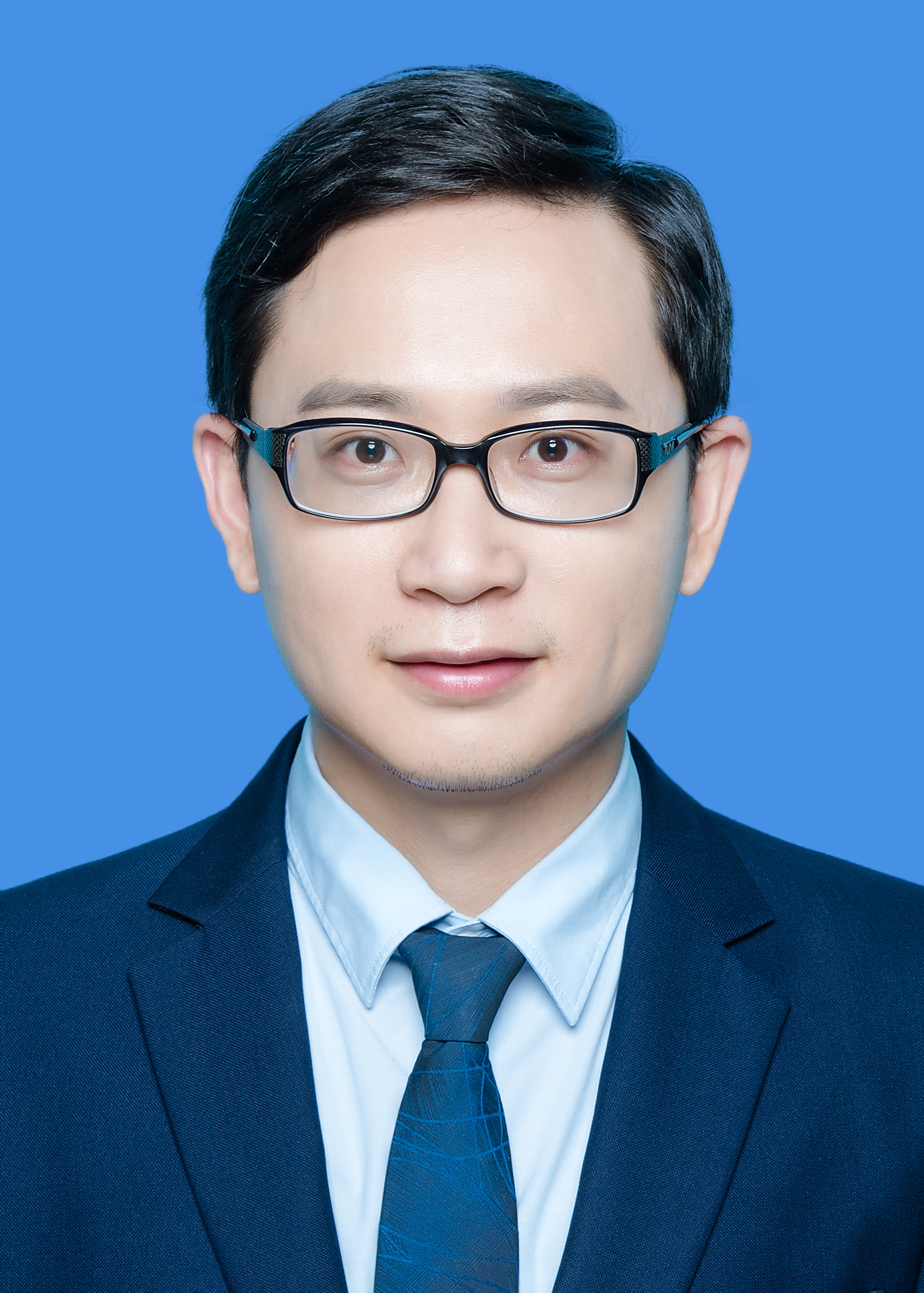}}]{Lingyan Ran} received B.S. and Ph.D. degrees from Northwestern Polytechnical University, Xi’an, China, in 2011 and 2018, respectively. From 2013 to 2015, he was a Visiting Scholar with the Stevens Institute of Technology, Hoboken, NJ, USA. He is currently an Associate Professor with the School of Computer Science, Northwestern Polytechnical University. His research interests include image classification and semantic segmentation. Dr. Ran is currently a member of CSIG.
\end{IEEEbiography}

\vspace{-25pt}

\begin{IEEEbiography}[{\includegraphics[width=1in,height=1.25in,clip,keepaspectratio]{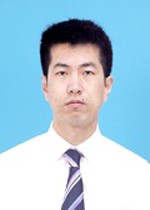}}]{Guoqiang Liang}
received a B.S. in automation and a Ph.D. degrees in pattern recognition and intelligent systems from Xi$'$an Jiaotong University (XJTU), Xi$'$an, China, in 2012 and 2018, respectively. From Mar. to Sep. 2017, he was a visiting Ph.D. Student with the University of South Carolina, Columbia, SC, USA. Currently, he is doing PostDoctoral Research at the School of Computer Science and Engineering, Northwestern Polytechnical University, Xi$'$an, China. His research interests include human pose estimation and human action classification.
\end{IEEEbiography}

\vspace{-25pt}

\begin{IEEEbiography}[{\includegraphics[width=1in,height=1.25in,clip,keepaspectratio]{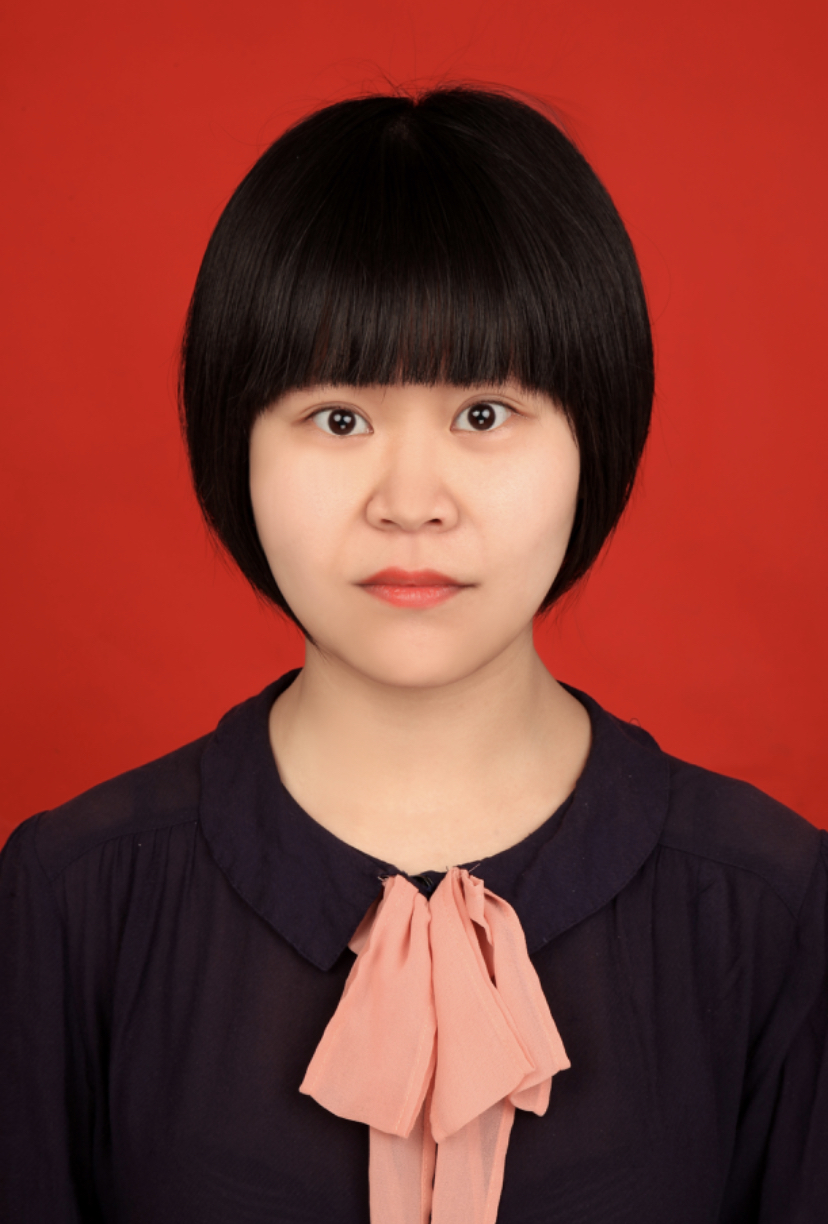}}]
{Hexu Wang} is currently an Associate Professor of Xijing University in China. 
She is also with the School of Information and Technology, Northwest University and Xi'an Key Laboratory of Human-Machine Integration and Control Technology for Intelligent Rehabilitation. Her main research interests focus on AI supply chain and machine learning.

\end{IEEEbiography}

\vspace{-25pt}

\begin{IEEEbiography}[{\includegraphics[width=1in,height=1.25in,clip,keepaspectratio]{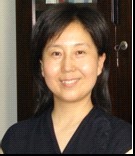}}]{Yanning Zhang}
(SM'10) received her B.S. degree from Dalian University of Science and Engineering in 1988, M.S. and Ph.D. degrees from Northwestern Polytechnical University in 1993 and 1996, respectively. She is presently a Professor at the School of Computer Science, Northwestern Polytechnical University. She is also the organization chair of the Ninth Asian Conference on Computer Vision (ACCV2009). Her research work focuses on signal and image processing, computer vision and pattern recognition. She has published over 200 papers in international journals, conferences and Chinese key journals.

\end{IEEEbiography}

\vfill

\end{document}

%% file: addon.tex
\newcommand{\vC}{{\bf C}}

\newcommand{\vF}{{\bf F}}

\newcommand{\vM}{{\bf M}}